\documentclass[a4paper]{article}
\usepackage[top=4cm, bottom=4cm, left=2cm, right=2cm]{geometry} 

\usepackage{hyperref}
\usepackage{amssymb}
\usepackage{amsthm}

\usepackage{amsmath}

\usepackage{float}
\usepackage{graphicx}

\usepackage{booktabs}
\usepackage{fancyhdr}
\usepackage{url}
\usepackage{xspace}
\usepackage{cleveref}
\usepackage{times}
\newcommand{\eprime}{{\sc Essence Prime}\xspace}
\newcommand{\savilerow}{{\sc Savile Row}\xspace}
\newcommand{\minion}{{\sc Minion}\xspace}
\newcommand{\essence}{{\sc Essence}\xspace}

\newcommand{\comment}[1]{}

\newcommand{\version}{1.10.1\xspace}

\begin{document}

\title{Savile Row Manual}

\author{Peter Nightingale}

\date{}

\maketitle

\begin{abstract}

We describe the constraint modelling tool Savile Row, its input language and its main features. Savile Row translates a solver-independent constraint modelling language to the input languages for various solvers including constraint, SAT, and SMT solvers. After a brief introduction, the manual describes the Essence Prime language, which is the input language of Savile Row. Then we describe the functions of the tool, its main features and options and how to install and use it.

\end{abstract}

\tableofcontents

\section{Introduction}

\savilerow is a constraint modelling tool. It provides a 
high-level language for the user to specify their constraint problem, and 
automatically translates that language to the input languages of constraint solvers
as well as SAT, MaxSAT, and SMT solvers. 
\savilerow is a research tool, designed to enable research into 
reformulation of constraint models, therefore it is designed for flexibility. 
It is designed to be easy to add new transformation rules and develop new translation pipelines. 

This manual is for version \version of \savilerow.

This manual covers the basics of using \savilerow. It does not cover adding
new rules or translation pipelines.

\savilerow reads the \eprime modelling language, which is described in \Cref{sec:eprime} below. \savilerow converts constraint models expressed in \eprime 
into the solver input format, in a process that has some similarities to 
compiling a high-level programming language. Like a compiler, \savilerow applies
some optimisations to the model (the most basic being partial evaluation and common subexpression elimination).

\subsection{Problem Classes and Instances}

The distinction between \textit{problem classes} and \textit{problem instances} 
will be important in this manual. It is easiest to start with an example. 
Sudoku in general is a problem class, and a particular Sudoku puzzle is an 
instance of that class. The problem class is described by writing down the 
rules of Sudoku, i.e. that we have a $9 \times 9$ grid and that each row, 
column and subsquare must contain all values $1..9$. A particular instance is 
described by taking the class and adding the clues, i.e.\ filling in some of 
the squares. The set of clues is a \textit{parameter} of the class -- in this
case the only parameter. Adding the parameters to a class to make a problem 
instance is called \textit{instantiation}. 

In typical use, \savilerow will read a problem class file and also a parameter file, 
both written in  \eprime. It will instantiate the problem class and unroll all quantifiers
and matrix comprehensions, then perform reformulations and flattening of nested expressions before 
producing output for a constraint solver. 

\subsection{Solver Backends}\label{sub:backends}

Output produced by \savilerow may be entirely flattened or it may contain some nested expressions depending on the backend solver or output language. Also, global constraints will be decomposed when not supported by the backend solver, another source of difference between the different backends. The SAT, MaxSAT, and SMT backends have a final encoding step to produce their lower-level representations. The backends are as follows:

\begin{itemize}
\item Minion -- Output is produced in the Minion 3 language for Minion 1.9.1 or later. 
The model produced is not entirely flat, it makes use of nested \texttt{watched-or} and \texttt{watched-and}
constraints.
\item Chuffed -- Output is produced for Chuffed in the entirely flat, instance-level language FlatZinc for use by the \texttt{fzn-chuffed} tool.
\item Gecode -- Similar to Chuffed with some differences in the set of constraints that are decomposed. 
\item OR-Tools -- Similar to Chuffed and Gecode, with some differences in the set of constraints that are decomposed. 
\item Standard FlatZinc -- Output for any solver with a FlatZinc parser. 
\item SAT -- Output is produced in DIMACS format for use with any SAT solver. An overview of the SAT encoding is given below. 
\item SMT -- Output is produced in SMT-LIB 2 format for use with SMT solvers such as Boolector, Z3, and Yices2.
\item MaxSAT -- Produces Weighted Partial MaxSAT for use with MaxSAT solvers such as Open-WBO and MaxHS. 
\item MiniZinc -- Output is produced in an instance-level, almost flat subset of the MiniZinc modelling language. 
The MiniZinc output follows the FlatZinc outputs as closely as possible. 
\end{itemize}

The Minion, Chuffed, OR-Tools, Gecode, standard FlatZinc, SAT, SMT, and MaxSAT backends are able to run the solver and parse the solution. 
For these backends, \savilerow will produce a solution file (in \eprime) and a
file with some statistics. The SAT and SMT backends make multiple calls to solve optimisation problems (using a bisecting strategy by default), unless the target solver natively supports optimisation. 
All constraints (i.e. the entire \eprime language) are supported with each backend. 
For the bundled solvers (Minion, Kissat (SAT), Chuffed), paths are set so that
\savilerow always uses the bundled solver by default. 

\minion is well supported by \savilerow so we will use \minion as the reference
in this document. Modelling problems directly in \minion's input
language is time-consuming and tedious because of its 
primitive structure (it can be compared to writing a 
complex program in assembly language), so a tool like \savilerow takes 
much of the tedium out of modelling constraint problems for \minion.

\section{The Essence Prime Language}\label{sec:eprime}

The purpose of this section is to describe the \eprime language and to be a 
reference for users of \eprime -- not to be a formal description of the langauge.   \eprime is a constraint modelling language,
therefore it is mainly designed for describing $\mathcal{NP}$-hard decision 
problems. It is not the only (or the first) constraint modelling language. 
\eprime began as a subset of \essence~\cite{essence-journal-08} and has been extended
from there. It is similar to the earlier Optimization Programming Language (OPL)~\cite{opl-book}.
\eprime is implemented by the tool \savilerow~\cite{nightingale2014automatically, nightingale2015automatically}.

\eprime is considerably different to procedural programming languages, it does not specify a procedure
to solve the problem. The user specifies the problem in terms of decision variables
and constraints, and the solver automatically finds a value for each variable 
such that all constraints are satisfied. This means, for example, that the 
order the constraints are presented in \eprime is irrelevant. 

\begin{figure}
\begin{center}
\includegraphics[width=0.3\textwidth]{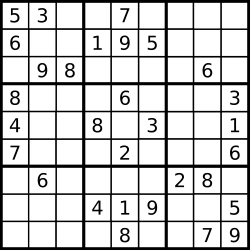}\:\:\includegraphics[width=0.3\textwidth]{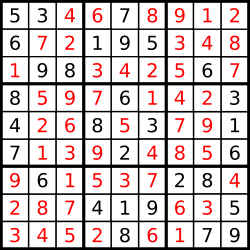}
\end{center}
\caption{\label{fig:sudoku}On the left is a Sudoku puzzle. The objective is to fill in all blank
squares using only the digits 1-9 such that each row contains all the digits 1-9, each column 
also contains all the digits 1-9, and each of the 3 by 3 subsquares outlined in thicker lines also 
contains all digits 1-9. 
On the right is the solution. 
(Images are public domain from Wikipedia.)
}
\end{figure}

\eprime allows the user to solve \textit{constraint satisfaction problems} (CSPs). 
A simple example of a CSP is a Sudoku puzzle (Figure~\ref{fig:sudoku}). To convert
a single Sudoku puzzle to CSP, each square can be represented as a decision
variable with domain $\{1\ldots 9\}$. The clues (filled in squares) and the rules of the puzzle are
easily expressed as constraints. 

We will use Sudoku as a running example. A simple first attempt of modelling Sudoku
in \eprime is shown in \Cref{sudoku1}. In this case the clues (for example \texttt{M[1,1]=5}) 
are included in the model. We have used not-equal constraints to state that all 
digits must be used at most once in each row and column. We have also omitted the sub-square
constraints for now. 

\begin{figure}
\begin{verbatim}
language ESSENCE' 1.0

find M : matrix indexed by [int(1..9), int(1..9)] of int(1..9)

such that

M[1,1]=5,
M[1,2]=3,
M[1,5]=7,
....
M[9,9]=9,

forAll row : int(1..9) . 
    forAll col1 : int(1..9) .
        forAll col2: int(col1+1..9) . M[row, col1]!=M[row, col2],

forAll col : int(1..9) . 
    forAll row1 : int(1..9) .
        forAll row2: int(row1+1..9) . M[row1, col]!=M[row2, col]
\end{verbatim}
\caption{First example of a (partial) Sudoku model in \eprime}\label{sudoku1}
\end{figure}

In this example, some CSP decision variables are declared using a \texttt{find} statement. 
It is also worth noting that other variables exist that are not decision variables,
for example, \texttt{row} is a \textit{quantifier} variable that exists only to apply some 
constraints to every row. 

An \eprime model usually describes a \textit{class} of CSPs. For example, it 
might describe the class of all Sudoku puzzles. In order to solve a particular
instance of Sudoku, the instance would be specified in a separate \textit{parameter} file (also 
written in \eprime). The model would have parameter variables (of type integer, boolean or matrix), 
and the parameter file would specify a value for each of these variables.

Since \eprime is a constraint modelling language, we will define the
constraint satisfaction problem (CSP). 
A CSP $\mathcal{P}=\langle\mathcal{X},\mathcal{D},\mathcal{C}\rangle$
is defined as a set of $n$ \textit{decision variables} 
$\mathcal{X}=\langle x_{1},\dots,x_{n}\rangle$,
a set of domains $\mathcal{D}=\langle D(x_{1}),\dots,D(x_{n})\rangle$ where
$D(x_{i})\subsetneq\mathbb{Z}$, $\left|D(x_{i})\right|<\infty$ is the
finite set of all potential values of $x_{i}$, and a conjunction
$\mathcal{C}=C_{1}\wedge C_{2}\wedge\cdots\wedge C_{e}$ of constraints. 

For CSP $\mathcal{P}=\langle\mathcal{X},\mathcal{D},\mathcal{C}\rangle$,
a constraint $C_{k}\in\mathcal{C}$ consists of a sequence of $m>0$
variables $\mathcal{X}_{k}=\langle x_{k_{1}},\dots,x_{k_{m}}\rangle$
with domains $\mathcal{D}_{k}=\langle D(x_{k_{1}}),\ldots,D(x_{k_{m}})\rangle$
s.t. $\mathcal{X}_{k}$ is a subsequence%
\footnote{I use subsequence in the sense that $\langle1,3\rangle$ is a subsequence 
of $\langle1,2,3,4\rangle$.%
} of $\mathcal{X}$, $\mathcal{D}_{k}$ is a subsequence of $\mathcal{D}$,
and each variable $x_{k_{i}}$ and domain $D(x_{k_{i}})$ matches a variable
$x_{j}$ and domain $D(x_{j})$ in $\mathcal{P}$. $C_{k}$ has an associated
set $C_{k}^{S}\subseteq D(x_{k_{1}})\times\dots\times D(x_{k_{m}})$ of
tuples which specify allowed combinations of values for the variables
in $\mathcal{X}_{k}$.

\subsection{The Essence Prime Expression Language}

In \eprime \textit{expressions} are built up from variables and literals using 
operators (such as \texttt{+}). We will start by describing the types
that expressions may take, then describe the simplest expressions and
build up from there. 

\subsubsection{Types and Domains}\label{sec:types-domains}

Types and domains play a similar role; they prescribe a set of
values that an expression or variable can take. Types denote non-empty sets that contain all elements
that have a similar structure, whereas domains denote possibly empty
sets drawn from a single type.  In this manner, each domain is associated with an
underlying type.  For example integer is the type underlying the
domain comprising integers between 1 and 10. The type contains all integers, 
and the domain is a finite subset. 

\eprime is a strongly typed language;
every expression has a type, and the types of all
expressions can be inferred and checked for correctness.
Furthermore, \eprime is a finite-domain language; every decision variable is
associated with a finite domain of values. 

The atomic types of \eprime are {\tt int} (integer) and {\tt bool} (boolean). 
There is also a compound type, \texttt{matrix},
that is constructed of atomic types.

There are three different types of domains in \eprime{}:
boolean, integer and matrix domains. Boolean and integer 
domains are both atomic domains; matrix domains are built from 
atomic domains.

\begin{description}

\item[Boolean Domains]
{\tt bool} is the Boolean domain consisting of {\tt false} and {\tt true}.

\item[Integer Domains]
An integer domain represents a finite subset of the integers, and is specified 
as a sequence of ranges or individual elements, for example 
{\tt int(1..3,5,7,9..20)}. Each range includes the endpoints, so the meaning of a range {\tt a..b} is the set 
\(\{ i \in \mathbb{Z} | a\le i \le b \}\). A range {\tt a..b} would normally be 
in order, i.e.~{\tt a<=b} but this is not strictly required. Out-of-order ranges 
correspond to the empty set of integers. 

The meaning of an integer domain is the union of the ranges and individual elements in the domain. 
For example, {\tt int(10, 1..5, 4..9)} is equivalent to {\tt int(1..10)}. 

Finally, the elements and endpoints of ranges may be expressions of type {\tt int}. 
The only restriction is that they may not contain any CSP decision variables.  
The integer expression language is described in the following sections.

\item[Matrix Domains]
A matrix is defined by the keyword {\tt matrix}, followed by its dimensions and 
the base domain. The dimensions are a list, in square brackets, of domains.
For instance,
\begin{verbatim}
Matrix1 : matrix indexed by [int(1..10),int(1..10)] of int(1..5) 
\end{verbatim}
is the domain of a two-dimensional square matrix, indexed by $1..10$ in both dimensions,
where each element of the matrix has the domain {\tt int(1..5)}. Elements of this matrix would 
be accessed by {\tt Matrix1[A,B]} where A and B are integer expressions. 

Matrices may be indexed by any integer domain or the boolean domain. For example, 

\begin{verbatim}
Matrix2 : matrix indexed by [int(-2..5),int(-10..10,15,17)] of int(1..5)
\end{verbatim}

is a valid matrix domain.

\end{description}

\subsubsection{Domain Expressions}\label{sec:domainexpressions}

\eprime contains a small expression language for boolean and integer domains. 
This language consists of three binary infix operators \texttt{intersect}, \texttt{union} and \texttt{-}.
All three are left-associative and the precedences are given in Appendix~\ref{app:op}.
The language also allows bracketed subexpressions with \texttt{(} and \texttt{)}. 

For example, the first and second lines below are exactly equivalent. 

\begin{verbatim}
letting dom be domain int(1..5, 3..8)
letting dom be domain int(1..5) union int(3..8)
\end{verbatim}

\subsubsection{Literals}\label{sec:literals}

Each of the three types (integer, boolean and matrix) has a corresponding literal 
syntax in \eprime. Any value of any type may be written as a literal. Sets and 
real numbers are not (as yet) part of the language.  Integer and boolean literals
are straightforward:

\begin{center}
\begin{verbatim}
1 2 3 -5
true false
\end{verbatim}
\end{center}

There are two forms of matrix literals. The simpler form is a comma-separated list of
expressions surrounded by square brackets. For example, the following is a matrix
literal containing four integer literals. 

\begin{center}
{\tt [ 1, 3, 2, 4 ]}
\end{center}

Matrix literals may contain any valid expression in \eprime. For example a
matrix literal is allowed to contain other matrix literals to build up a matrix 
with two or more dimensions. The types of the expressions contained in the matrix
literal must all be the same. 

The second form of matrix literal has an explicit index domain that specifies how 
the literal is indexed. This is specified after the comma-separated list of
contents using a \texttt{;} as follows.

\begin{center}
{\tt [ 1, 3, 2, 4 ; int(4..6,8) ]}
\end{center}

The index domain must be a domain of type \texttt{bool} or \texttt{int}, and must 
contain the same number of values as the number of items in the matrix literal. 
When no index domain is specified, a matrix of size \texttt{n} is indexed by \texttt{int(1..n)}. 

Finally a multi-dimensional matrix value can be expressed by nesting matrix literals.
Suppose we have the following domain:

\begin{center}
\texttt{matrix indexed by [int(-2..0),int(1,2,4)] of int(1..5)}
\end{center}

One value contained in this domain is the following:

\begin{verbatim}
[ [ 1,2,3 ; int(1,2,4) ],
  [ 1,3,2 ; int(1,2,4) ],
  [ 3,2,1 ; int(1,2,4) ]
  ; int(-2..0) ]
\end{verbatim}

\subsubsection{Variables}\label{sec:vars}

Variables are identified by a string. Variable names must start with a letter 
\texttt{a-z} or \texttt{A-Z}, and after the first letter may contain any of
the following characters: \texttt{a-z A-Z 0-9 \_}. A variable may be of type
integer, boolean or matrix. 

Scoping of variables depends on how they are declared and is dealt with in the
relevant sections below. As well as a type, variables have a category. The 
category is \textit{decision}, \textit{quantifier} or \textit{parameter}. Decision
variables are CSP variables, and the other categories are described below. 

Expressions containing decision variables are referred to as \textit{decision expressions},
and expressions containing no decision variables as \textit{non-decision expressions}.
This distinction is important because expressions in certain contexts are not allowed
to contain decision variables. 

\subsubsection{Expression Types}

Expressions can be of any of the three basic types (integer, boolean or matrix). 
Integer expressions range over an integer domain, 
for instance {\tt x + 3} (where \texttt{x} is an integer variable) is an integer expression ranging from 
$lb(x)+3$ to $ub(x)+3$. Boolean expressions range over the 
Boolean domain, for instance the integer comparison 
 {\tt x = 3} can either be {\tt true}
or {\tt false}. 

\subsubsection{Type Conversion}

Boolean expressions or literals are automatically converted to integers when used
in a context that expects an integer. As is conventional \texttt{false} is converted
to \texttt{0} and \texttt{true} is converted to {1}. For example, the following are 
valid \eprime boolean expressions. 

\begin{verbatim}
1+2-3+true-(x<y)=5
false < true
\end{verbatim}

Integer expressions are not automatically converted to boolean. Matrix expressions
cannot be converted to any other type (or vice versa). 

\subsubsection{Matrix Indexing and Slicing}\label{sub:matrix-slicing}

Suppose we have a three-dimensional matrix {\tt M} with the following domain:

\begin{verbatim}
matrix indexed by [int(1..3), int(1..3), bool] of int(1..5)
\end{verbatim}

{\tt M} would be indexed as follows: {\tt M[x,y,z]}, where \texttt{x} and \texttt{y} may be integer or
boolean expressions and \texttt{z} must be boolean. Because the matrix has the base
domain \texttt{int(1..5)}, {\tt M[x,y,z]} will be an integer expression. 
Matrix indexing is a partial function: when one of the indices is out of bounds 
then the expression is undefined. \eprime has the relational semantics, 
in brief this means that the boolean expression containing the undefined expression is
\texttt{false}. So for example, \texttt{M[1,1,false]=M[2,4,true]} is always \texttt{false}
because the 4 is out of bounds. The relational semantics are more fully described 
in Section~\ref{sec:undef} below. 

Parts of matrices can be extracted by \textit{slicing}. Suppose we have the 
following two-dimensional matrix named \texttt{N}: 

\begin{verbatim}
[ [ 1,2,3 ; int(1,2,4) ],
  [ 1,3,2 ; int(1,2,4) ],
  [ 3,2,1 ; int(1,2,4) ]
  ; int(-2..0) ]
\end{verbatim}

We could obtain the first row by writing \texttt{N[-2,..]}, which is equal to \texttt{[ 1,2,3 ; int(1..3)]}. 
Similarly the first column can be obtained by 
\texttt{N[..,1]} which is \texttt{[ 1,1,3 ; int(1..3)]}.  In general, the indices in a matrix slice may be \texttt{..} or 
an integer or boolean expression that does not contain any decision variables. 
Matrix slices are always indexed contiguously from 1 regardless of the original matrix
domain. 

When a matrix slice has an integer or boolean index that is out of bounds, then
the expression is undefined and this is handled as described in Section~\ref{sec:undef}.

\subsubsection{Integer and Boolean Expressions}\label{sec:int-bool-expressions}

\eprime has a range of binary infix and unary operators and functions for building up 
integer and boolean expressions, for example:

\begin{itemize}
\item Integer operators: {\tt + - * ** / \% | min max}
\item Boolean operators: {\tt \verb1\/ /\1 -> <-> !}
\item Quantified sum: {\tt sum}
\item Logical quantifiers: {\tt forAll exists}
\item Numerical comparison operators: {\tt = != > < >= <=}
\item Matrix comparison operators: {\tt <=lex <lex >=lex >lex}
\item Set operator: {\tt in}
\item Global constraints: {\tt allDiff gcc cumulative}
\item Table constraint: {\tt table}
\end{itemize}

These are described in the following subsections. 

\subsubsection{Integer Operators}

\eprime has the following binary integer 
operators: {\tt + - * / \% **}. {\tt +}, {\tt -} and {\tt *} are the standard
integer operators. 

The operators \texttt{/} and \texttt{\%} are integer division and modulo functions. 
{\tt a/b} is defined as $\lfloor a/b \rfloor$ (i.e. it always rounds down). 
This does not match some programming languages, for example C99 which rounds 
towards 0. 

The modulo operator {\tt a\%b} is defined as $a-b\lfloor a/b \rfloor$ to be 
complementary to \texttt{/}.

\begin{verbatim}
3/2 = 1
(-3)/2 = -2
3/(-2) = -2
(-3)/(-2) = 1

3 % 2 = 1
(-3) % 2 = 1
3 % (-2) = -1
(-3) % (-2) = -1
\end{verbatim}

{\tt **} is the power function: {\tt x**y} is defined as $x^y$.
There are two unary functions: absolute value (where {\tt |x|} is the absolute 
value of {\tt x}), and negation (unary {\tt -}).

\subsubsection{Boolean Operators}

\eprime has the {\tt \verb1/\1} (and) and {\tt \verb1\/1} (or) operators 
defined on boolean expressions. There are also {\tt ->} (implies), {\tt <->} 
(if and only if), and {\tt !} (negation). These operators all take boolean 
expressions and produce a new boolean expression. They can be nested arbitrarily.  

The comma {\tt ,} in \eprime is also a binary boolean operator, with the same meaning 
as {\tt \verb1/\1}. However it has a different precedence, and is used quite 
differently. {\tt \verb1/\1} is normally used within a constraint, and
{\tt ,} is used to separate constraints (or separate groups of constraints 
constructed using a \texttt{forAll}). Consider the following example.

\begin{verbatim}
forAll i : int(1..n) . x[i]=y[i] /\ x[i]!=y[i+1],
exists i : int(1..n) . x[i]=1 /\ y[i]!=y[i+1]
\end{verbatim}

Here we have two quantifiers, both with an {\tt \verb1/\1} inside. The comma is 
used to separate the {\tt forAll} and the {\tt exists}. The comma has the lowest precedence of all binary operators.

\subsubsection{Integer and Boolean Functions}

\eprime has named functions {\tt min(X,Y)} and {\tt max(X,Y)} that both take 
two integer expressions \texttt{X} and \texttt{Y}. 
\texttt{min} and \texttt{max} can also be applied to one-dimensional matrices to
obtain the minimum or maximum of all elements in the matrix (see Section~\ref{sec:matrix-functions}).
\texttt{factorial(x)} returns the factorial of values from 0 to 20 where the result fits in
a 64-bit signed integer. It is undefined for other values of \texttt{x} and the expression \texttt{x} is not 
allowed to contain decision variables. \texttt{popcount(x)} returns the bit count of the 
64-bit two's complement representation of \texttt{x}, and \texttt{x} may not contain 
decision variables. 

The function \texttt{toInt(x)} takes a boolean expression \texttt{x} and 
converts to an integer 0 or 1. This function is included only for compatibility with 
\textsc{Essence}: it is not needed in \eprime because booleans are automatically cast to
integers. 

\subsubsection{Numerical Comparison Operators}

\eprime provides the following integer comparisons with their obvious meanings: 
{\tt = != > < >= <=}. These operators each take two integer expressions and produce
a boolean expression. 

\subsubsection{Matrix Comparison Operators}

\eprime provides a way of comparing one-dimensional matrices. These operators compare two 
matrices using the dictionary order (\textit{lexicographical} order, or lex for 
short). 

There are four operators. {\tt A <lex B} ensures that A comes before B in 
lex order, and {\tt A <=lex B} which ensures that {\tt A <lex B} or {\tt A=B}.
{\tt >=lex} and {\tt >lex} are identical but with the arguments reversed. 
For all four operators, A and B may have different lengths and may be indexed differently, but they must be
one-dimensional.  Multi-dimensional matrices may be flattened to one dimension using
the \texttt{flatten} function described in Section~\ref{sec:matrix-functions} below. 

\subsubsection{Set Operator}

The operator \texttt{in} states that an integer expression takes a value in a 
set expression. The set espression may not contain decision variables. 
The set may be a domain expression 
(Section~\ref{sec:domainexpressions}) or the \texttt{toSet} function that
converts a one-dimensional matrix to a set, as in the examples below. 

\begin{verbatim}
x+y in (int(1,3,5) intersect int(3..10))
x+y in toSet([ i | i : int(1..n), i%2=0])
\end{verbatim}

\subsubsection{The Quantified Sum Operator}\label{sub:qsum}

The {\tt sum} operator corresponds to the mathematical 
$\sum$ and has the following syntax:
\begin{center}
{\tt sum i : D . E} 
\end{center}
where {\tt i} is a quantifier variable, {\tt D} is a domain, and {\tt E} is the
expression contained within the {\tt sum}. More than one quantifier variable may
be created by writing a comma-separated list {\tt i,j,k}. 

For example, if we want to take the sum of all numbers in the range 1 to 10 we write

\begin{verbatim}
sum i : int(1..n) . i
\end{verbatim}

which corresponds to $\sum\nolimits_{i=1}^n i$. {\tt n} cannot be a decision 
variable.

Quantified sum has several similarities to the {\tt forAll} and {\tt exists} 
quantifiers (described below in Section~\ref{sub:forallexists}): it introduces
new local variables (named quantifier variables) that can be used within {\tt E}, 
and the quantifier variables all have the same domain {\tt D}. However {\tt sum} 
has one important difference: a {\tt sum} is an integer expression. 

A quantified sum may be nested inside any other integer operator, including another
quantified sum:

\begin{verbatim}
sum i,j : int(1..10) .
    sum k : int(i..10) .
        x[i,j] * k
\end{verbatim}

\subsubsection{Universal and Existential Quantification} \label{sub:forallexists}

Universal and existential quantification are powerful means to 
write down a series of constraints in a compact way. 
Quantifications have the same syntax as {\tt sum}, but with 
{\tt forAll} and {\tt exists} as keywords:

\begin{verbatim}
forAll  i : D . E
exists  i : D . E
\end{verbatim}

For instance, the universal quantification

\begin{verbatim}
forAll i : int(1..3) . x[i] = i
\end{verbatim}

corresponds to the conjunction:

\begin{verbatim}
x[1] = 1 /\ x[2] = 2 /\ x[3] = 3
\end{verbatim}

An example of existential quantification is

\begin{verbatim}
exists i : int(1..3) . x[i] = i
\end{verbatim}

and it corresponds to the following disjunction:

\begin{verbatim}
x[1] = 1 \/ x[2] = 2 \/ x[3] = 3
\end{verbatim}

Quantifications can range over several quantified variables and can 
be arbitrarily nested, as demonstrated with the {\tt sum} quantifier.

In the running Sudoku example, \texttt{forAll} quantification is used to build
the set of constraints. The expression:

\begin{verbatim}
forAll row : int(1..9) . 
    forAll col1 : int(1..9) .
        forAll col2: int(col1+1..9) . M[row, col1]!=M[row, col2]
\end{verbatim}

is a typical use of universal quantification. 

\subsubsection{Quantification over Matrix Domains}

All three quantifiers are defined on matrix domains as well as integer and 
boolean domains. For example, to quantify over all permutations of size \texttt{n}:

\begin{verbatim}
forAll perm : matrix indexed by [int(1..n)] of int(1..n) . 
    allDiff(perm) -> exp
\end{verbatim}

The variable \texttt{perm} represents a matrix drawn from the matrix domain, and the 
\texttt{allDiff} constraint evaluates to \texttt{true} when \texttt{perm} is a 
permutation of $1\ldots n$. Hence the expression \texttt{exp} is quantified for
all permutations of $1\ldots n$. 

If \texttt{n} is a constant, the example above could be written as a set of \texttt{n}
nested \texttt{forAll} quantifiers. However if \texttt{n} is a parameter of the 
problem class, it is very difficult to write the example above using other (non-matrix) 
quantifiers. 

\subsubsection{Global Constraints}

\eprime provides a small set of global constraints such as {\tt allDiff} (which is satisfied when a
vector of variables each take different values). Global constraints are all
boolean expressions in \eprime. 
Typically it is worth using these in models because the solver often performs
better with a global constraint compared to a set of simpler constraints. 

For example, the following two lines are semantically equivalent (assuming
{\tt x} is a matrix indexed by {\tt 1..n}). 

\begin{verbatim}
forAll i,j : int(1..n) . i<j -> x[i]!=x[j]
allDiff(x)
\end{verbatim}

Both lines will ensure that the variables in {\tt x} take different values. 
However the {\tt allDiff} will perform better in most situations.\footnote{In the 
current version of Savile Row, with default settings, the \texttt{x[i]!=x[j]} constraints would be aggregated to create \texttt{allDiff(x)} therefore there is no difference in performance between these two statements. 
There would be a difference in performance when aggregation is switched off (for example by using the \texttt{-O1} flag).}
Table \ref{tab:globals} summarises the global constraints available in \eprime.

To give another example, the \texttt{cumulative} constraint for scheduling 
applies a bound on the amount of a resource that may be used at each time step in a schedule.
It has arguments \texttt{X}, \texttt{Dur}, \texttt{Res}, and \texttt{Bound} (see Table \ref{tab:globals})
representing the start time of each task, the duration of each task, the resource
required for each task while it is running, and the upper limit on the resource usage. 
Suppose there are three tasks, each task takes 3 time units to execute, each task 
consumes 2 units of resource when it is running, and the resource limit is 5:

\begin{verbatim}
Res=[2,2,2]
Bound=5
Dur=[3,3,3]
\end{verbatim}

If all three tasks are running at any time step, they would exceed the resource bound. 
\texttt{X=[0, 1, 2]} is not a solution, but \texttt{X=[0, 1, 3]} is (the first task finishes before the third task starts).

\begin{table}  
    \begin{center}
    \begin{tabular}{p{0.25\textwidth}p{0.2\textwidth}p{0.45\textwidth}}\toprule 
Global Constraint & Arguments & Description \\
\hline
{\tt allDiff(X)} & {\tt X} is a matrix & Ensures expressions in {\tt X} take distinct values in any solution. \\
\hline
{\tt atleast(X, C, Vals)} & {\tt X} is a matrix, {\tt Vals} is a matrix of non-decision expressions, {\tt C} is a matrix of non-decision expressions & For each non-decision expression {\tt Vals[i]}, the number of occurrences of {\tt Vals[i]} in {\tt X} is at least {\tt C[i]}. \\
\hline
{\tt atmost(X, C, Vals)} & {\tt X} is a matrix, {\tt Vals} is a matrix of non-decision expressions, {\tt C} is a matrix of non-decision expressions & For each non-decision expression {\tt Vals[i]}, the number of occurrences of {\tt Vals[i]} in {\tt X} is at most {\tt C[i]}. \\
\hline
{\tt gcc(X, Vals, C)} & {\tt X} is a matrix, {\tt Vals} is a matrix of non-decision expressions, {\tt C} is a matrix & For each non-decision expression {\tt Vals[i]}, the number of occurrences of {\tt Vals[i]} in {\tt X} equals {\tt C[i]}.\\
\hline
{\tt alldifferent\verb1_1except\newline\quad(X, Value)} & {\tt X} is a matrix, {\tt Value} is a non-decision expression & Ensures variables in {\tt X} take distinct values, except that {\tt Value} may occur any number of times.\\ 
\hline
{\tt cumulative\newline\quad(X, Dur, Res, Bound)} & \texttt{X}, \texttt{Dur}, and \texttt{Res} are matrices, \texttt{Bound} is a scalar. & The cumulative constraint enforces a resource limit on a set of tasks. The start time of a task \texttt{i} is \texttt{X[i]}, the duration of \texttt{i} is \texttt{Dur[i]}, and the resource required by task \texttt{i} is \texttt{Res[i]}. At each time step, the sum of the resource in use (i.e.\ the sum of the resource required by tasks that are running at that time step) cannot be more than \texttt{Bound}.\\
\bottomrule
      \end{tabular}
\end{center}
\caption{Global constraints in \eprime. Each may be nested within expressions and have arbitrary expressions
    nested within them. Each matrix parameter of a global constraint must be a one-dimensional
    matrix. In some cases the parameter is a matrix of \textit{non-decision expressions} -- that is, integer or 
    boolean expressions that contain no decision variables. Quantifier and parameter variables are allowed
    in non-decision expressions. }
\label{tab:globals}
\end{table}

Now we have the \texttt{allDiff} global constraint, the Sudoku example can be improved
and simplified as shown in \Cref{sudoku2}.

\begin{figure}
\begin{verbatim}
language ESSENCE' 1.0

find M : matrix indexed by [int(1..9), int(1..9)] of int(1..9)

such that

M[1,1]=5,
M[1,2]=3,
M[1,5]=7,
....
M[9,9]=9,

forAll row : int(1..9) .
    allDiff(M[row,..]),

forAll col : int(1..9) .
    allDiff(M[..,col])	 
\end{verbatim}
\caption{\label{sudoku2}Second version of Sudoku, using global constraints.}
\end{figure}

Global constraints are boolean expressions like any other, and are allowed to be
used in any context that accepts a boolean expression. 

\subsubsection{Table Constraints}

In a table constraint the satisfying tuples of the constraint are specified using a matrix. 
This allows a table constraint to theoretically implement any relation, although
practically it is limited to relations where the set of satisfying tuples is small
enough to store in memory and efficiently search. 

The first argument specifies the variables in the scope of the
constraint, and the second argument is a two-dimensional matrix of satisfying tuples. 
For example, the constraint {\tt a+b=c} on boolean variables could be written
as a table as follows. 

\begin{verbatim}
table( [a,b,c], [[0,0,0], [0,1,1], [1,0,1]] )
\end{verbatim}

The first argument of \texttt{table} is a one-dimensional matrix expression. It
may contain both decision variables and constants. The second argument is a 
two-dimensional matrix expression containing no decision variables. The second argument
can be stated as a matrix literal, or an identifier, or by slicing a higher-dimension
matrix, or by constructing a matrix using matrix comprehensions (see Section~\ref{sec:comprehensions}). 

If the same matrix of tuples is used for many table constraints, a {\tt letting} 
statement can be used to state the matrix once and use it many times. Lettings are
described in Section~\ref{sec:letting} below. 

\subsubsection{Matrix Comprehensions}\label{sec:comprehensions}

We have seen that matrices may be written explicitly as a matrix literal (Section~\ref{sec:literals}), and 
that existing matrices can be indexed and sliced (Section~\ref{sub:matrix-slicing}).  Matrices
can also be constructed using \textit{matrix comprehensions}. This provides a 
very flexible way to create matrices of variables or values. A single 
matrix comprehension creates a one-dimensional matrix, however they can be
nested to create multi-dimensional matrices. There are two syntactic forms of matrix
comprehension:

\begin{verbatim}
[ exp | i : domain1, j : domain2, cond1, cond2 ]
[ exp | i : domain1, j : domain2, cond1, cond2 ; indexdomain ]
\end{verbatim}

where \texttt{exp} is any integer, boolean or matrix expression. 
This is followed by any number of comprehension variables, each with a domain.
After the comprehension variables we have an optional list of conditions: these
are boolean expressions that constrain the values of the comprehension variables. 
Finally there is an optional index domain. This provides an index domain to the 
constructed matrix. 

To expand the comprehension, each assignment to the comprehension 
variables that satisfies the conditions is enumerated in lexicographic order. 
For each such assignment, the values of the comprehension variables are substituted into \texttt{exp}.
The resulting expression then becomes one element of the constructed matrix.

The simplest matrix comprehensions have only one comprehension variable, as in the example
below. 

\begin{verbatim}
[ num**2 | num : int(1..5) ] = [ 1,4,9,16,25 ; int(1..5) ]
\end{verbatim}

The matrix constructed by a comprehension is one-dimensional and is either indexed
from 1 contiguously, or has the given index domain. The given domain must have a lower bound
but is allowed to have no upper bound. 
For example the first line below produces the matrix on the second line. 

\begin{verbatim}
[ i+j | i: int(1..3), j : int(1..3), i<j ; int(7..) ]
[ 3, 4, 5 ; int(7..9) ]
\end{verbatim}

Matrix comprehensions allow more advanced forms of slicing than the matrix slice syntax in
Section~\ref{sub:matrix-slicing}. For example it is possible to slice an arbitrary subset of the rows or 
columns of a two-dimensional matrix. The following two nested comprehensions will 
slice out the entries of a matrix \texttt{M} where both rows and columns are odd-numbered, 
and build a new two-dimensional matrix. 

\begin{verbatim}
[ [ M[i,j] | j : int(1..n), j%2=1 ] | i : int(1..n), i%2=1 ]
\end{verbatim}

Now we have matrix comprehensions, the Sudoku example can be completed by adding
the constraints on the $3\times 3$ subsquares (as shown in \Cref{sudoku3}). A comprehension is used to 
create a matrix of variables and the matrix is used as the parameter of an
\texttt{allDiff} constraint. 

\begin{figure}[tbp]
\begin{verbatim}
language ESSENCE' 1.0

find M : matrix indexed by [int(1..9), int(1..9)] of int(1..9)

such that

M[1,1]=5,
M[1,2]=3,
M[1,5]=7,
....
M[9,9]=9,

forAll row : int(1..9) .
    allDiff(M[row,..]),

forAll col : int(1..9) .
    allDiff(M[..,col]),	 

$ all 3x3 subsquare have to be all-different
$ i,j indicate the top-left corner of the subsquare. 
forAll i,j : int(1,4,7) .
    allDiff([ M[k,l]  | k : int(i..i+2), l : int(j..j+2)])
\end{verbatim}
\caption{Third version of Sudoku, using global constraints, a matrix comprehension and quantifiers.}\label{sudoku3}
\end{figure}

In this example, the matrix constructed by the comprehension depends on the values 
of \texttt{i} and \texttt{j} from the \texttt{forAll} quantifier.  The comprehension 
variables \texttt{k} and \texttt{l} each take one of three values, to cover the 9 entries \texttt{M[k,l]} in the 
subsquare. 

\subsubsection{Matrix Comprehensions over Matrix Domains}

Similarly to quantifiers, matrix comprehension variables can have a matrix domain.
For example, the following comprehension builds a two-dimensional matrix where
the rows are all permutations of \texttt{1..n}. 

\begin{verbatim}
[ perm | perm : matrix indexed by [int(1..n)] of int(1..n), allDiff(perm) ]
\end{verbatim}

\subsubsection{Functions on Matrices\label{sec:matrix-functions}}

Table \ref{tab:matfunc} lists the matrix functions available in \eprime. 

\begin{table}
\centering
 \begin{tabular}{lp{0.2\textwidth}p{0.5\textwidth}}\toprule 
Function & Arguments & Description \\
\hline
{\tt sum(X)} & {\tt X} is a one-dimensional matrix & Constructs the sum of elements in \texttt{X} \\
\hline
{\tt product(X)} & {\tt X} is a one-dimensional matrix & Constructs the product of elements in \texttt{X} \\
\hline
{\tt and(X)} & {\tt X} is a one-dimensional matrix of booleans & Constructs the conjunction of \texttt{X} \\
\hline
{\tt or(X)} & {\tt X} is a one-dimensional matrix of booleans & Constructs the disjunction of \texttt{X} \\
\hline
{\tt min(X)} & {\tt X} is a one-dimensional matrix & The integer minimum of elements in \texttt{X} \\
\hline
{\tt max(X)} & {\tt X} is a one-dimensional matrix & The integer maximum of elements in \texttt{X} \\
\hline
{\tt flatten(X)} & {\tt X} is a matrix & Constructs a one-dimensional matrix (indexed contiguously from 1) with the same contents as \texttt{X} \\
\hline
{\tt flatten(n,X)} & {\tt X} is a matrix & The first \texttt{n+1} dimensions of \texttt{X} are flattened into one dimension that is indexed contiguously from 1. Therefore \texttt{flatten(n,X)} produces a new matrix with \texttt{n} fewer dimensions than \texttt{X}. The first argument \texttt{n} must be positive. \\
\hline
{\tt cat(A,B,C,...)} & {\tt A}, {\tt B}, {\tt C} etc are matrices, each with the same number of dimensions. & Concatenate the given matrices into a single matrix by combining their first dimension. For example, if \texttt{A=[[1,2],[3,4]]} and \texttt{B=[[5,6]]}, then \texttt{cat(A,B)=[[1,2],[3,4],[5,6]]}. The first dimension of the resulting matrix is always indexed contiguously from 1. \\
\hline
{\tt list(A,B,C,...)} & {\tt A}, {\tt B}, {\tt C} etc are scalars or matrices of type integer or Boolean. & Constructs a one-dimensional matrix (indexed contiguously from 1) with the same contents as \texttt{A,B,C,...}. Each matrix in the list of arguments \texttt{A,B,C,...} is flattened (as in \texttt{flatten}) and each scalar becomes a one-dimensional matrix with a single element. The resulting matrices are concatenated into a single one-dimensional matrix (as in \texttt{cat}). \\
\hline
{\tt toSet(X)} & {\tt X} is a one-dimensional matrix of non-decision expressions & The set of elements in \texttt{X} \\
\bottomrule
\end{tabular}
\caption{Matrix Functions\label{tab:matfunc}}
\end{table}

The functions \texttt{sum}, \texttt{product}, \texttt{and} and \texttt{or} were originally
intended to be used with matrix comprehensions, but can be used with any matrix. 
The quantifiers \texttt{sum}, \texttt{forAll} and \texttt{exists}
can be replaced with \texttt{sum}, \texttt{and} and \texttt{or} containing matrix comprehensions. 
For example, consider the \texttt{forAll} expression below (from the Sudoku model). It can be replaced with
the second line below. 

\begin{verbatim}
forAll row : int(1..9) . allDiff(M[row,..])

and([ allDiff(M[row,..]) | row : int(1..9) ])
\end{verbatim}

In fact, matrix functions combined with matrix comprehensions are strictly more
powerful than quantifiers. Also, the function \texttt{product} has no corresponding
quantifier. Quantifiers are retained in the language because they can be easier to read.

As a more advanced example, given an $n\times n$ matrix \texttt{M} of decision variables, the sum below
is the determinant of \texttt{M} using the Leibniz formula. The outermost comprehension
constructs all permutations of $1\ldots n$ using a matrix domain and an \texttt{allDiff}.
Lines 3 and 4 contain a comprehension that is used to obtain the number of \textit{inversions}
of \texttt{perm} (the number of pairs of values that are not in ascending order). 
Finally line 5 builds a product of some of the entries of the matrix. 
Without the \texttt{product} function, it is difficult (perhaps impossible) to write the Leibniz formula
in \eprime for a matrix of unknown size $n$. 

\begin{verbatim}
sum([
    $  calculate the sign of perm from the number of inversions. 
    ( (-1) ** sum([ perm[idx1]>perm[idx2] 
                  | idx1 : int(1..n), idx2 : int(1..n), idx1<idx2 ]) )*   
    product([ M[j, perm[j]] | j : int(1..n) ])
| perm : matrix indexed by [int(1..n)] of int(1..n), allDiff(perm)])
\end{verbatim}

The \texttt{flatten} function is typically used to feed the contents of a 
multi-dimensional matrix expression into a constraint that accepts only 
one-dimensional matrices. For example, given a three-dimensional matrix \texttt{M},
the following example is a typical use of \texttt{flatten}.  

\begin{verbatim}
allDiff( flatten(M[1,..,..]) )
\end{verbatim}

When flattening a matrix \texttt{M} to create a new matrix \texttt{F}, the order
of elements in \texttt{F} is defined as follows. Suppose \texttt{M} were written as
a matrix literal (as in Section~\ref{sec:literals}) the order elements are written
in the matrix literal is the order the elements appear in \texttt{F}. The following
example illustrates this for a three-dimensional matrix. 

\begin{verbatim}
flatten([ [ [1,2], [3,4] ], [ [5,6], [7,8] ] ]) = [1,2,3,4,5,6,7,8]
\end{verbatim}

\subsection{Model Structure}

An \eprime model is structured in the following way:

\begin{enumerate}
\item Header with version number: {\tt  language ESSENCE' 1.0} 
\item Parameter declarations (optional)
\item Constant definitions (optional)
\item Decision variable declarations (optional)
\item Constraints on parameter values (optional)
\item Objective (optional)
\item Solver Control (optional)
\item Constraints
\end{enumerate}

Parameter declarations, constant definitions, decision variable 
declarations, and constraints on parameter values can be interleaved, but for readability we suggest to put them in the 
order given above. Comments are preceded by `\$' and run to the end of the line.

Parameter values are defined in a separate file, the 
{\em parameter file}. Parameter files have the same header 
as problem models and hold a list of parameter definitions.
Table \ref{tab:modelstructure} gives an overview of the model
structure of problem and parameter files.
Each model part will be discussed in more detail in the following sections.

\begin{table}
\begin{center}
\begin{tabular}{ll}
\toprule
Problem Model Structure & Parameter File Structure \\
\hline
{\tt language ESSENCE' 1.0}  & {\tt language ESSENCE' 1.0}  \\
\ & \\
\$ {\em parameter declaration}       & \$ {\em parameter instantiation} \\
{\tt given n : int}         & {\tt letting n=7} \\
\$ {\em constant definition}        & \\
{\tt letting c=5 }  & \\
\ & \\
\$ {\em variable declaration }       & \\
{\tt find x,y : int(1..n) } & \\
\ & \\
\$ {\em constraints}                 & \\
{\tt such that } & \\
\ \ {\tt x + y >= c,}    & \\  
\ \ {\tt x + c*y = 0}    & \\  
\bottomrule
\end{tabular}
\caption{Model Structure of problem files and parameter files in \eprime. `\$' denotes comments.}\label{tab:modelstructure}
\end{center}
\end{table}

\subsubsection{Parameter Declarations with {\tt given}}

Parameters are declared with the {\tt given} keyword followed 
by a domain the parameter ranges over. Parameters are allowed to 
range over the infinite domain {\tt int}, or domains that contain an open range
such as {\tt int(1..)} and {\tt int(..10)}. For example,

\begin{verbatim}
given n : int(0..)

given d : int(0..n)
\end{verbatim}

declares two parameters, and the domain of the second depends on the value of
the first. Parameters may have integer, boolean or matrix domains. 

Now we have parameters we can generalise the Sudoku model of \Cref{sudoku2} to represent the 
problem class of all Sudoku puzzles. The generalised model is shown in \Cref{sudoku4}. The parameter is the \texttt{clues} 
matrix, where blank spaces are represented as \texttt{0} and non-zero entries
in \texttt{clues} are copied to \texttt{M}. 

\begin{figure}
\begin{verbatim}
language ESSENCE' 1.0

given clues : matrix indexed by [int(1..9), int(1..9)] of int(0..9)

find M : matrix indexed by [int(1..9), int(1..9)] of int(1..9)

such that

forAll row : int(1..9) .
    forAll col : int(1..9) .
        (clues[row, col]!=0) -> (M[row, col]=clues[row, col]),

forAll row : int(1..9) .
    allDiff(M[row,..]),

forAll col : int(1..9) .
    allDiff(M[..,col]),	 

$ all 3x3 subsquare have to be all-different
$ i,j indicate the top-left corner of the subsquare. 
forAll i,j : int(1,4,7) .
    allDiff([ M[k,l]  | k : int(i..i+2), l : int(j..j+2)])
\end{verbatim}
\caption{Fourth version of Sudoku, adding the \texttt{clues} parameter to the version in \Cref{sudoku3}.}\label{sudoku4}
\end{figure}

\subsubsection{Constant Definitions with {\tt letting}\label{sec:letting}}

In most problem models there are re-occurring constant values and
it can be useful to define them as constants. The {\tt letting}
statement allows to assign a name with a constant value. The statement

\begin{verbatim}
letting NAME = A
\end{verbatim}

introduces a new identifier {\tt NAME} that is associated with 
the expression \texttt{A}. Every subsequent occurrence of 
{\tt NAME} in the model is replaced by the value of {\tt A}. Please note 
that {\tt NAME} cannot be used in the model {\em before} it has been 
defined. \texttt{A} may be any integer, boolean or matrix expression that does
not contain decision variables. Some integer examples are shown below. 

\begin{verbatim}
given n : int(0..)
letting c = 10
letting d = c*n*2
\end{verbatim}

Here the second integer constant depends on the first. As well as integer and
boolean expressions, lettings may contain matrix expressions, as in the
example below. When using a matrix literal the domain is optional -- the two 
lettings below are equivalent.  The version with the matrix domain may be useful when
the matrix is not indexed from 1. 

\begin{verbatim}
letting cmatrix = [ [2,8,5,1], [3,7,9,4] ]
letting cmatrix2 : matrix indexed by [ int(1..2), int(1..4) ] of int(1..10) 
    = [ [2,8,5,1], [3,7,9,4] ]
\end{verbatim}

Finally new matrices may be constructed using a slice or comprehension, as in the example below
where the letting is used for the table of a table constraint. 

\begin{verbatim}
letting xor_table = [ [a,b,c] | a : bool, b : bool, c : bool, 
                                (a /\ b) \/ (!a /\ !b) <-> c ]
find x,y,z : bool
such that
table([x,y,z], xor_table)
\end{verbatim}

\subsubsection{Constant Domains}

Constant domains are defined in a similar way using the keywords {\tt be domain}.

\begin{verbatim}
letting c = 10
given n : int(1..)
letting INDEX be domain int(1..c*n)
\end{verbatim}

In this example {\tt INDEX} is defined to be an integer domain, the upper bound of
which depends on a parameter \texttt{n} and another constant \texttt{c}. 

Constant domains are convenient when a domain is reused several times. In the 
Sudoku running example, we could use a letting for the domain \texttt{int(1..9)}, as shown in \Cref{sudoku5}.

\begin{figure}
\begin{verbatim}
language ESSENCE' 1.0
letting range be domain int(1..9)

given clues : matrix indexed by [range, range] of int(0..9)

find M : matrix indexed by [range, range] of range

such that

forAll row : range .
    forAll col : range .
        (clues[row, col]!=0) -> (M[row, col]=clues[row, col]),

forAll row : range .
    allDiff(M[row,..]),

forAll col : range .
    allDiff(M[..,col]),	 

$ all 3x3 subsquare have to be all-different
$ i,j indicate the top-left corner of the subsquare. 
forAll i,j : int(1,4,7) .
    allDiff([ M[k,l]  | k : int(i..i+2), l : int(j..j+2)])
\end{verbatim}
\caption{Fifth version of the Sudoku model. Compared to \Cref{sudoku4} the \texttt{range} letting statement has been added.}\label{sudoku5}
\end{figure}

\subsubsection{Decision Variable Declaration with {\tt find}}

Decision variables are declared using the {\tt find} keyword followed by a name and their
corresponding domain. The domain must be finite. The example below
\begin{verbatim}
find x : int(1..10)
\end{verbatim}
defines a single decision variable with the given domain. 
It is possible to define several variables in one \texttt{find} by giving multiple
names, as follows. 

\begin{verbatim}
find x,y,z : int(1..10)
\end{verbatim}

Matrices of decision variables are declared using a matrix domain, as in the following example. 

\begin{verbatim}
find m : matrix indexed by [ int(1..10) ] of bool
\end{verbatim}

This declares \texttt{m} as a 1-dimensional matrix of 10 boolean variables. Simple and matrix
domains are described in Section~\ref{sec:types-domains}.

In the Sudoku running example, we have been using the following two-dimensional matrix domain.

\begin{verbatim}
find M : matrix indexed by [int(1..9), int(1..9)] of int(1..9)
\end{verbatim}

\subsection{Constraints on Parameters with {\tt where}}

In some cases it is useful to restrict the values of the parameters. This is
achieved with the \texttt{where} keyword, which is followed by a boolean expression
containing no decision variables. In the following example, we require the first
parameter to be less than the second. 

\begin{verbatim}
given x : int(1..)
given y : int(1..)
where x<y
\end{verbatim}

\subsection{Objective}
The objective of a problem is either {\tt maximising} or 
{\tt minimising} an integer or boolean expression. For instance,

\begin{verbatim}
minimising x
\end{verbatim}

states that the value assigned to variable $x$ will be minimised.
Only one objective is allowed, and it is placed after all \texttt{given}, 
{\tt find} and {\tt letting} statements.

\subsection{Solver Control}

In addition to instructing the solver to minimise or maximise some expression, 
\eprime also allows a variable order to be specified for search. 
The \texttt{branching on} statement specifies a sequence of variables for the 
solver to branch on. Each element of the \texttt{branching on} list may be an individual variable or a matrix with any number of dimensions. Decision variables in matrices are enumerated in 
the order produced by the \texttt{flatten} function. Any matrix expression is allowed within the \texttt{branching on} list, including matrix comprehensions. A simple use of \texttt{branching on} would be to list some individual variables:

\begin{verbatim}
find w,x,y,z : int(1..10)

branching on [w,x]
\end{verbatim}

The \texttt{branching on} list may contain the same decision variable
more than once. This can be useful to pick some variables from a matrix to
branch on first, then include the rest of the matrix simply by including the entire matrix.
In the following example the diagonal of \texttt{M} is branched first, then the rest
of \texttt{M} is included. 

\begin{verbatim}
find M : matrix indexed by [int(1..9), int(1..9)] of int(1..9)

branching on [ [ M[i,i] | i : int(1..9)], M ]
\end{verbatim}

In some cases a good search order does not follow the natural order of matrices, and might even alternate between matrices. The following example uses a comprehension to alternate between two matrices that are both indexed by timestep:

\begin{verbatim}
find arm : matrix indexed by [int(1..timesteps)] of int(1..10)
find action : matrix indexed by [int(1..timesteps)] of int(1..10)

branching on [ [ [arm[i], action[i]] | i : int(1..timesteps) ] ]
\end{verbatim}

Some solver types (e.g.\ SAT, MaxSAT, and SMT solvers) ignore the search order, while Chuffed (with its free search option) uses the search order as part of a hybrid search.

When searching for all solutions (\texttt{-all-solutions}) or a given number of solutions (\texttt{-num-solutions}) with a \texttt{branching on} statement, each solution produced will have a distinct assignment to the \texttt{branching on} variables, thus using a \texttt{branching on} statement can reduce the number of solutions found. 

When using Minion as the backend solver, a \texttt{heuristic} statement may be used in the preamble to specify a dynamic search heuristic 
to apply to the variables in the \texttt{branching on} list. \texttt{heuristic} 
is followed by \texttt{static} (the default), \texttt{sdf}, \texttt{conflict} or \texttt{srf} (refer to the Minion documentation for a description of these heuristics).  If any other solver is used then the \texttt{heuristic} statement is ignored.

%

\subsection{Constraints}

After defining constants and declaring decision variables and parameters,
constraints are specified with the keywords {\tt such that}. The constraints in
\eprime are boolean expressions as described in Section~\ref{sec:int-bool-expressions}.

Typically the constraints are written as a list of boolean expressions separated
by the \texttt{,} operator.  


\section{Undefinedness in \eprime}\label{sec:undef}

Since the current version of \eprime is a closed language, there are a finite set
of partial functions in the language. For example, \texttt{x/y} is a partial function
because it is not defined when \texttt{y=0}. In its current version \eprime implements the 
relational semantics as defined by Frisch and Stuckey \cite{frisch-stuckey-undef}.
The relational semantics has the advantage that it can be implemented efficiently. 

The relational semantics may be summarised as follows:
\begin{itemize}
\item Any integer or matrix expression directly containing an undefined expression is itself undefined.
\item Any domain or domain expression directly containing an undefined expression is itself undefined.
\item Any statement in the preamble (\texttt{find}, \texttt{letting} etc) directly containing an undefined expression is undefined. 
\item Any boolean expression that directly contains an undefined expression is \texttt{false}. 
\end{itemize}

Informally, the relational semantics confines the effect of an undefined expression
to a small part of the problem instance (which becomes \texttt{false}), in many cases avoiding
making the entire problem instance \texttt{false}.

Consider the four examples below. Each contains a division by zero which is an undefined
integer expression. In each case the division is contained in a comparison. Integer
comparisons are boolean expressions. 

\begin{verbatim}
(x/0 = y) = false
(x/0 != y) = false
! (x/0 = y) = true
! (x/0 != y) = true
\end{verbatim}

Applying the rules of the relational semantics results in each of the comparisons inside the brackets 
becoming \texttt{false}:

\begin{verbatim}
(false) = false
(false) = false
! (false) = true
! (false) = true
\end{verbatim}

In the relational semantics, \texttt{(x/0 != y)} is not semantically equivalent to \texttt{!(x/0 = y)},
which is somewhat counter-intuitive. 

Another counter-intuitive case arises with matrix indexing. In the following example,
the expression \texttt{M[0]} is undefined because 0 is not in the index domain. 
If the matrix is boolean (i.e.\ \texttt{DOM} is \texttt{bool}) then \texttt{M[0]}
becomes \texttt{false}, and the model has a solution when \texttt{M[1]=false}.
However, if the matrix contains integer variables (i.e.\ \texttt{DOM} is \texttt{int(0..1)})
then the constraint \texttt{M[0] = M[1]} becomes \texttt{false} and the model has no solutions. 

\begin{verbatim}
find M : matrix indexed by [int(1)] of DOM
such that
M[0] = M[1]
\end{verbatim}

In the \savilerow implementation of \eprime, all partial functions are removed
in a two-step process, before any other transformations are applied. The first step
is as follows. For each partial function a boolean 
expression is created that is \texttt{true} when the partial function is
defined and \texttt{false} when it is undefined. There are six operators that may be
partial: division, modulo, power, factorial, matrix indexing and matrix slicing. 
Table~\ref{tab:undef} shows the generated boolean expression for each operator. 
The boolean expression is then added to the model by connecting it (with \verb1/\1) onto the closest boolean
expression above the partial function in the abstract syntax tree.

\begin{table}  
    \begin{center}
    \begin{tabular}{ll}\toprule 
Partial Function & Defined When \\
\hline
{\tt X/Y} & \texttt{Y!=0} \\
\hline
{\tt X\%Y} & \texttt{Y!=0} \\
\hline
{\tt X**Y} & \verb1(X!=0 \/ Y!=0) /\ Y>=01 \\
\hline
\texttt{factorial(X)} & \texttt{X>=0} \\
\hline
Matrix indexing:   &  $\mathtt{I}\in\mathtt{D1}$ \verb1/\1  \\
\texttt{M[I,J,K]}  &  $\mathtt{J}\in\mathtt{D2}$ \verb1/\1  \\
where domain of \texttt{M} is  &  $\mathtt{K}\in\mathtt{D3}$  \\
\texttt{matrix indexed by [D1, D2, D3] of DBase}   &  \\
\hline
Matrix slicing:   &  $\mathtt{I}\in\mathtt{D1}$ \verb1/\1  \\
\texttt{M[I,..,K]}  &   $\mathtt{K}\in\mathtt{D3}$  \\
where domain of \texttt{M} is &    \\
\texttt{matrix indexed by [D1, D2, D3] of DBase}  &  \\
\bottomrule
\end{tabular}
\end{center}
    \caption{Partial functions in \eprime. \texttt{X}, \texttt{Y}, \texttt{I}, \texttt{J} and \texttt{K} are arbitrary expressions
    of the correct type.}
\label{tab:undef}
\end{table}

The second step is to replace the partial function \texttt{P} with a total function \texttt{SP}.
For each input where \texttt{P} is defined, \texttt{SP} is defined to the same value. For inputs where \texttt{P} is 
undefined, \texttt{SP} takes a default value (typically \texttt{0} for integer expressions).

Once both steps have been applied to each partial function, the model is well
defined everywhere. This is done first, before any other model transformations,
and thus allows all subsequent transformations to be simpler because there is no
need to allow for undefinedness. 
For example, the expression \texttt{x/y=x/y} may not be simplified to \texttt{true},
because it is \texttt{false} when \texttt{y=0}. However, after replacing the partial
division function, the resulting equality can be simplified to \texttt{true}.  In
general any equality between two syntactically identical expressions can be simplified
to \texttt{true} once there are no partial functions.


\section{Installing and Running \savilerow}

\savilerow is distributed as an archive with the following contents: 

\begin{itemize}
\item The Java source code (in {\tt src}) licensed with GPL 3.
\item The compiled classes in a JAR file named {\tt savilerow.jar}.
\item Executable script {\tt savilerow} for running \savilerow.
\item This document in {\tt doc}.
\item Various example \eprime files and parameter files, in {\tt examples}.
\item Required Java libraries in {\tt lib}
\item Executables for backend solvers Minion, Kissat, and Chuffed, and the graph automorphism solver Ferret in {\tt bin}
\item Build script \texttt{compile.sh} to rebuild \texttt{savilerow.jar}\footnote{Requires a JDK to be installed and \texttt{javac} to be in the path. It compiles for Java 8 backward compatibility by using the \texttt{--release 8} flag. If compiling using Java 8 the \texttt{--release 8} flag needs to be removed from the \texttt{javac} command-line.}. 
\end{itemize}

Archives are provided for Linux and Mac, and the Linux archive works well on 
Windows 10 in the Ubuntu app. The two distributions are largely the same,
with the only difference being the packaged executables in {\tt bin}.

A recent version of Java is required on all platforms. \savilerow was
compiled to run on Java 8 or later. 

\subsection{Running \savilerow on Linux and Mac}

Download the appropriate archive and unpack it somewhere convenient. Open a 
terminal and navigate to the \savilerow directory.  
Use the script named \texttt{savilerow}. One of
the simplest ways of running \savilerow is given below.

\begin{verbatim}
./savilerow problemFile.eprime parameterFile.param
\end{verbatim}

The first argument is the problem class file. This 
is a plain text file containing the constraint problem, expressed in the \eprime
language. 

The second argument ({\tt parameterFile.param}) is the parameter file (again in the 
\eprime language). This contains the data for the problem instance. 
The parameter file can be omitted if the problem has no parameters (no \texttt{given} statements in 
the preamble). 

\subsection{Running \savilerow on Windows}

The current (\version) version of \savilerow does not have a release for 
Windows specifically, however the Linux release works well in the 
Ubuntu 20.04 app on Windows 10. To install, first enable the Windows Subsystem for 
Linux. Second, install the Ubuntu 20.04 app and update packages in it (for example with 
\texttt{sudo apt update; sudo apt upgrade}). Third, install Java (for example, 
\texttt{sudo apt install default-jre}). Then follow the instructions for installing 
and running on Linux.

\subsection{Solution Files}

For the Minion, SAT, SMT, MaxSAT, Standard FlatZinc, Gecode, and Chuffed backends, \savilerow is able to run the solver and parse the solution (or set of solutions) 
produced by the solver. These solutions are translated back into \eprime.  
For each \texttt{find} statement in the model file (i.e.\ each statement that declares decision variables), 
the solution file contains
a matching \texttt{letting} statement. For example, if the model file contains the following \texttt{find} statement:

\begin{verbatim}
find M : matrix indexed by [int(1..2), int(1..2)] of int(1..5)
\end{verbatim}

The solution file could contain the following \texttt{letting} statement. 

\begin{verbatim}
letting M = [[2,3],
             [1,2]]
\end{verbatim}

\subsection{File Names}\label{sub:filenames}

The input files for \savilerow typically have the extension \texttt{.eprime}, \texttt{.param}, or \texttt{.eprime-param}.
These extensions allow \savilerow to identify the model and parameter file on the 
command line. If these files have a different extension (or no extension) then
\savilerow must be called in a slightly different way:

\begin{verbatim}
./savilerow -in-eprime problemFile -in-param parameterFile
\end{verbatim}

Given input file names \texttt{<problemFile>} and \texttt{<parameterFile>}, output files have the following
names by default.

\begin{itemize}
\item For Minion output, \texttt{<parameterFile>.minion} (or if there is no parameter file, 
\texttt{<problemFile>.minion}).
\item For Gecode or Chuffed output, \texttt{<parameterFile>.fzn} (or if there is no parameter file, 
\texttt{<problemFile>.fzn}).
\item For SAT or MaxSAT output, \texttt{<parameterFile>.dimacs} (or if there is no parameter file, 
\texttt{<problemFile>.dimacs}).
\item For Minizinc output, \texttt{<parameterFile>.mzn} (or if there is no parameter file, 
\texttt{<problemFile>.mzn}).
\end{itemize}

\begin{sloppypar}
When \savilerow parses a single solution from the output of a solver, it produces a file named
\texttt{<parameterFile>.solution} (or if there is no parameter file, 
\texttt{<problemFile>.solution}). When there are multiple solutions (e.g.\ when using
the \texttt{-all-solutions} flag) the solution files are numbered (for example, 
nurses.param.solution.000001 to nurses.param.solution.000871). 

If \savilerow runs Minion or a SAT solver it produces a file \texttt{<parameterFile>.info} (or if there is no parameter file, 
\texttt{<problemFile>.info}) containing solver statistics.

Finally, a file named \texttt{<parameterFile>.aux}
(or \texttt{<problemFile>.aux}) is also created. This contains the symbol table
and is read if \savilerow is called a second time to parse a solution. 
\end{sloppypar}

\subsection{Command Line Options}

The command-line help text is produced with the following flag.

\begin{verbatim}
-help
\end{verbatim}

\subsubsection{Specifying input files}

The options \texttt{-in-eprime} and \texttt{-in-param} specify the input files
as in the example below. 

\begin{verbatim}
savilerow -in-eprime sonet.eprime -in-param sonet1.param
\end{verbatim}

These flags may be omitted for any filename that ends with \texttt{.eprime}, \texttt{.param}, or \texttt{.eprime-param}. 

The option \texttt{-params} may be used to specify the parameters on the command
line. For example, suppose the \texttt{nurse.eprime} model has two parameters. We 
can specify them on the command line as follows. The format of the parameter string
is identical to the format of a parameter file (where, incidentally, the \texttt{language} line is optional and line breaks are not significant).

{\footnotesize
\begin{verbatim}
savilerow nurse.eprime -params "letting n_nurses=4 letting Demand=[[1,0,1,0],[0,2,1,0]]"
\end{verbatim}}

\subsubsection{Specifying output format}

\savilerow produces output for various solvers and classes of solver as 
described in \Cref{sub:backends}. The 
output format is specified by using one of the following command-line options. 

\begin{verbatim}
-minion
-gecode
-chuffed
-or-tools
-flatzinc
-minizinc
-sat
-smt
-maxsat
\end{verbatim}

The output filename may be specified as follows. In each case there is a default
filename so the flag is optional. Default filenames are described in \Cref{sub:filenames}.
The flag \texttt{-out-sat} also sets the MaxSAT filename. 

\begin{verbatim}
-out-minion <filename>
-out-flatzinc <filename>
-out-minizinc <filename>
-out-sat <filename>
-out-smt <filename>
\end{verbatim}

The flag \texttt{-out-flatzinc} sets the FlatZinc filename.  

In addition, the file names for solution files, solver statistics files and 
symbol table files (called aux files) may be 
specified as follows. Once again there are default filenames described in Section~\ref{sub:filenames}.

\begin{verbatim}
-out-solution <filename>
-out-info <filename>
-out-aux <filename>
\end{verbatim}

The following flag sets each of the output files to \texttt{prefix.extension} where \texttt{extension} is the conventional one for the file type (i.e.\ \texttt{.minion}, \texttt{.fzn}, \texttt{.dimacs}, etc). 

\begin{verbatim}
-out-prefix <prefix>
\end{verbatim}

The symbol table is not saved by default. To obtain the symbol table file, use the following flag. The symbol table is used in ReadSolution mode, described below. 

\begin{verbatim}
-save-symbols
\end{verbatim}

\subsubsection{Optimisation Levels}

The optimisation levels (\verb|-O0| to \verb|-O3|) provide an easy way to control
how much optimisation \savilerow does, without having to switch on or off individual
optimisations. The default is \verb|-O2|, which is intended to provide a generally
recommended set of optimisations. The rightmost \texttt{-O} flag on the command line
is the one that takes precedence. 

\begin{verbatim}
-O0
\end{verbatim}

The lowest optimisation level, \texttt{-O0}, turns off all optional optimisations.
\savilerow will still simplify expressions (including constraints). Any expression containing
only constants will be replaced with its value. Some expressions have quite sophisticated 
simplifiers that will run even at \texttt{-O0}. For example, \texttt{allDiff([x+y+z, z+y+x, p, q])}
would simplify to \texttt{false} because the first two expressions in the \texttt{allDiff} are
symbolically equal after normalisation. 

\texttt{-O0} will do no common subexpression 
elimination, will not unify equal variables, and will not filter the domains of variables. 

\begin{verbatim}
-O1
\end{verbatim}

\texttt{-O1} does optimisations that are very efficient in both space and time. 
Variables that are equal are unified, and a form of common subexpression elimination
is applied (Active CSE, described below). \texttt{-O1} is equivalent to
\texttt{-deletevars} and \texttt{-active-cse}. 

\begin{verbatim}
-O2
\end{verbatim}

In addition to the optimisations performed by \texttt{-O1}, \texttt{-O2} performs
filtering of variable domains (both \texttt{find} and auxiliary variables) and aggregation (both of which are described in the 
following section).  \texttt{-O2} is the default optimisation level. \texttt{-O2} adds \texttt{-reduce-domains-extend} and \texttt{-aggregate} compared to \texttt{-O1}. 

\begin{verbatim}
-O3
\end{verbatim}

In addition to \texttt{-O2}, \texttt{-O3} enables tabulation (\texttt{-tabulate}) and 
associative-commutative common subexpression elimination (\texttt{-ac-cse}) (both described below). 

\subsubsection{Symmetry-Breaking Levels}

The options \verb|-S0|, \verb|-S1| and \verb|-S2| control optional optimisations
that can change the number of solutions. The default is \texttt{-S1} which applies
only very simple and fast optimisations. 

\begin{verbatim}
-S0
\end{verbatim}

Preserve the number of solutions, i.e.\ perform no optimisations that might
change the number of solutions.

\begin{verbatim}
-S1
\end{verbatim}

Remove variables that appear in no constraints, and allow \savilerow to introduce
auxiliary variables that are not functionally defined by the primary (`find') variables. 
At present, non-functional auxiliary variables are only used with the SAT, 
MaxSAT and SMT backends. Equivalent to \texttt{-remove-redundant-vars} and \texttt{-aux-non-functional}. 

\begin{verbatim}
-S2
\end{verbatim}

In addition to \texttt{-S1}, apply a graph automorphism solver to find symmetries
among the decision variables. The symmetries are then broken using the standard
lex-leader method. Equivalent to adding the flag \texttt{-var-sym-breaking} to \texttt{-S1}. 

\subsubsection{Translation Options}

\subsubsection*{Common Subexpression Elimination}

\savilerow currently implements four types of common subexpression elimination (CSE). 
Identical CSE finds and eliminates syntactically identical expressions. This
is the simplest form of CSE, however it can be an effective optimisation.
Active CSE performs some reformulations (for example applying De Morgan's laws)
to reveal expressions that are semantically equivalent but not syntactically 
identical. Active CSE subsumes Identical CSE. Active CSE is enabled by default as part of \texttt{-O2}.
Associative-Commutative CSE (AC-CSE) works on the associative and commutative (AC) operators
\texttt{+}, \texttt{*}, \verb1/\1 (and) and \verb1\/1 (or). It is able to
rearrange the AC expressions to reveal common subexpressions among them. 
AC-CSE is not enabled by default. It would normally be used in conjunction with
Identical or Active CSE. 
Finally, Active AC-CSE combines one active reformulation (integer negation) with
AC-CSE, so for example Active AC-CSE is able to extract $y-z$ from the three expressions
$x+y-z$, $w-x-y+z$, and $10-y+z$, even though the sub-expression occurs as $y-z$ in one
and $-y+z$ in the other two. Active AC-CSE is identical to AC-CSE for And, Or and
Product, it differs only on sum. 

The following flags control CSE. The first, \texttt{-no-cse}, turns off all CSE. 
The other flags each turn on one type of CSE. 

\begin{verbatim}
-no-cse
-identical-cse
-active-cse
-ac-cse
-active-ac-cse
\end{verbatim}

\subsubsection*{Variable Deletion}

\savilerow can remove a decision variable (either variables declared with \texttt{find} or
auxiliary variables introduced during tailoring) when the variable is equal to a 
constant, or equal to another variable. This is often a useful optimisation. It can be
enabled using the following flag. 

\begin{verbatim}
-deletevars
\end{verbatim}

\subsubsection*{Domain Filtering}

It can be useful to filter the domains of variables. In \savilerow this is done by
running the translation pipeline to completion and producing a Minion file, then 
running Minion usually with the options \texttt{-preprocess SACBounds\textunderscore{}limit -outputCompressedDomains}.
With these options Minion performs conventional propagation plus SAC on the 
lower and upper bound of each variable (SACBounds, also known as shaving) with a limit on the 
number of iterations. The filtered
domains are then read back in to \savilerow. The translation process is started again
at the beginning.  Thus domain filtering can benefit the entire translation process:
variables can be removed (with \texttt{-deletevars}), constraint expressions can be
simplified or even removed (if they are entailed), the number of auxiliary variables may be reduced.
In some cases domain filtering can enable another optimisation, for example on
the BIBD problem it enables associative-commutative CSE to do some very effective
reformulation. 

Domain filtering can be used with any target solver (but Minion
is always used to perform the domain filtering regardless of the target solver). It is 
switched on using the following flag. 

\begin{verbatim}
-reduce-domains
\end{verbatim}

Standard domain filtering affects the decision variables defined by \texttt{find} statements.
Auxiliary variables created by \savilerow are not filtered by default. To enable 
filtering of both \texttt{find} variables and auxiliary variables, use the following flag:

\begin{verbatim}
-reduce-domains-extend
\end{verbatim}

\subsubsection*{Aggregation of Constraints}

Aggregation collects sets of constraints together to form global constraints
that typically propagate better in the target solver. At present there are two
aggregators for allDifferent and GCC. \savilerow
constructs allDifferent constraints by collecting not-equal, less-than and 
other shorter allDifferent constraints. GCC is aggregated by collecting \texttt{atmost} and
\texttt{atleast} constraints over the same scope. 

\begin{verbatim}
-aggregate
\end{verbatim}

\subsubsection*{Tabulation}

Tabulation converts constraint expressions into table constraints to
improve propagation.  A set of four heuristics are used to identify candidate
expressions for tabulation.  When tabulating a constraint, the number of tuples
is limited to 10,000 to avoid spending excessive time on one constraint. Refer to 
the CP 2018 paper for more detail~\cite{tabulation-cp18}. 

\begin{verbatim}
-tabulate
\end{verbatim}

\subsubsection*{Factor Encoding}

The factor encoding of table constraints \cite{factor-encoding-2014} may be useful to strengthen propagation when a problem class has overlapping table constraints (or tabulation produces overlapping table constraints). It is enabled with the following flag. 

\begin{verbatim}
-factor-encoding
\end{verbatim}

\subsubsection*{Integer Variable Types}

When creating Minion output \savilerow will by default use the \texttt{DISCRETE} or \texttt{BOOL} variable type
when the variable has fewer than 10,000 values and \texttt{BOUND} otherwise. 
\texttt{DISCRETE} and \texttt{BOOL} represent the entire domain and \texttt{BOUND} only stores the
upper and lower bound, thus propagation may be lost when using \texttt{BOUND}. 
The following flag prevents \savilerow using \texttt{BOUND} variables. 

\begin{verbatim}
-no-bound-vars
\end{verbatim}

\subsubsection*{Remove Redundant Variables}

The following flag causes \savilerow to remove any decision variables that are
not mentioned in a constraint or in the objective function. It is enabled by 
default as part of \texttt{-S1}.

\begin{verbatim}
-remove-redundant-vars
\end{verbatim}

\subsubsection*{Non-functional Auxiliary Variables}

The following flag allows \savilerow to to create auxiliary variables
that are not functionally defined on the 
primary variables. May change the number of
solutions. 
Currently only affects the SAT, SMT, MaxSAT backends. Typically a non-functional variable would be \(\leq\) or \(\geq\) an expression rather than equal to the expression, thus avoiding encoding one side of the equality. 

\begin{verbatim}
-aux-non-functional
\end{verbatim}

\subsubsection*{Variable Symmetry Breaking}

If variable symmetry breaking is enabled, \savilerow will apply a graph 
automorphism solver based on Ferret to find symmetries among the decision variables.   
Any symmetries discovered will be broken by the standard lex-leader method.  
It is disabled by default because it can be expensive. Variable 
symmetry breaking is part of \texttt{-S2}.

\begin{verbatim}
-var-sym-breaking
\end{verbatim}

\subsubsection{SAT Encoding Options}\label{sec:sat-options}

The SAT backend in general is described in \Cref{app:satenc}. There are several options for encoding constraints into SAT, in particular sums and tables. 

Sum constraints are divided into three groups and there are flags to control the encoding of each group. The simplest group is at-most-one (AMO) and exactly-one (EO) constraints containing only Boolean terms (not just Boolean variables but any expression of type Boolean) and where all coefficients are 1. The encoding of these constraints is controlled by \texttt{-sat-amo} flags. Pseudo-Boolean (PB) sums also contain only Boolean terms but can have coefficients other than 1. The encoding of PB constraints is controlled by \texttt{-sat-pb} flags. Finally, the encoding of general sums (containing at least one integer term) is controlled by \texttt{-sat-sum} flags. 

\subsubsection*{Table Encodings}

Binary table constraints use the support encoding. For non-binary tables, the default encoding introduces a SAT variable for each tuple and was proposed by Bacchus~\cite{bacchus2007gac}. 
Another encoding based on MDDs is available for non-binary tables, enabled with the following flag:

\begin{verbatim}
-sat-table-mdd
\end{verbatim}

\subsubsection*{AMO Encodings}

The default encoding for AMO and EO constraints is Chen's 2-product, it is small and tends to be efficient. Other encodings are available: the Commander encoding of Klieber and Kwon (with group size 3); the ladder encoding; an encoding similar to the totalizer (named Tree); the pairwise encoding (with no auxiliary SAT variables); and the bimander encoding with group size 2. 
The flags to select the encoding are below:

\begin{verbatim}
-sat-amo-product  (default)
-sat-amo-commander
-sat-amo-ladder
-sat-amo-tree
-sat-amo-pairwise
-sat-amo-bimander
\end{verbatim}

\subsubsection*{PB Encodings}

There are nine encodings of Pseudo-Boolean constraints. Eight of them are based on Bofill et al~\cite{bofillPBAMO2022} (with thanks to Jordi Coll for providing the implementations): Multi-valued Decision Diagrams (MDD); the Global Polynomial Watchdog (GPW); the Local Polynomial Watchdog (LPW); the Sequential Weighted Counter (SWC); the Generalized Generalized Totalizer (GGT); the Reduced Generalized Generalized Totalizer (RGGT); another version of GGT with a tree-construction heuristic (GGTH); and the Generalized n-Level Modulo Totalizer (GMTO). Each of these is implemented exactly as described in~\cite{bofillPBAMO2022}. For these eight encodings, any pseudo-Boolean equality constraints are decomposed into \(\leq\) and \(\geq\). Also these eight encodings support only top-level constraints (i.e.\ constraints that are not nested inside a logic expression). The flags are as follows:

\begin{verbatim}
-sat-pb-mdd
-sat-pb-gpw
-sat-pb-lpw
-sat-pb-swc
-sat-pb-ggt
-sat-pb-rggt
-sat-pb-ggth
-sat-pb-gmto
\end{verbatim}

The final encoding (called Tree) is similar to the Generalized Totalizer. It can directly encode sum-equality, and can also encode non-top-level constraints. It is the default encoding and also the fallback for non-top-level constraints when using one of the other encodings. 

\begin{verbatim}
-sat-pb-tree  (default)
\end{verbatim}

The first eight encodings (MDD, GPW, LPW, SWC, GGT, RGGT, GGTH, GMTO) are able to take advantage of at-most-one (and exactly-one) relations on the Boolean variables to reduce the size of the encoding. All four of these encodings work with automatic AMO and EO detection as described in the CP 2019 paper~\cite{amopb-cp19}, which uses Minion to find mutexes between variables then constructs maximal cliques of the mutex graph. The flag below is used to switch on the AMO and EO detection. It is often worthwhile for sufficiently challenging instances with PB constraints. 

\begin{verbatim}
-amo-detect
\end{verbatim}

\subsubsection*{Sum Encodings}

The options for encoding general sums (any sum containing at least one integer variable) are the same as for PB constraints. The flags are as follows:

\begin{verbatim}
-sat-sum-mdd
-sat-sum-gpw
-sat-sum-lpw
-sat-sum-swc
-sat-sum-ggt
-sat-sum-rggt
-sat-sum-ggth
-sat-sum-gmto
-sat-sum-tree  (default)
\end{verbatim}

The Tree encoding natively handles integer terms. The other eight (MDD, GPW, LPW, SWC, GGT, RGGT, GGTH, GMTO) treat integer terms as a group of Boolean terms (e.g. \(1*(x=1) + 2*(x=2) + \cdots\)) with an exactly-one relation on them. These eight encodings also can benefit from automatically detected AMO and EO relations with the \texttt{-amo-detect} flag. 

\subsubsection{SMT Encoding Options}\label{sec:smt-options}

The SMT backend, enabled with the \texttt{-smt} flag, is able to produce the SMT-LIB 2 language with a choice of four theories: QF\_BV (the theory of bit vectors), QF\_IDL (integer difference logic), QF\_LIA (linear integer arithmetic), and QF\_NIA (nonlinear integer arithmetic). The theory is selected using one of the four flags below:

\begin{verbatim}
-smt-bv  (default)
-smt-idl
-smt-lia
-smt-nia
\end{verbatim}

For each theory there is a choice of \textit{nested} or \textit{flat} encodings, where nested aims to maintain the original nesting of expressions (for example, a product within a sum) as far as possible, whereas flat removes nesting by introducing auxiliary variables. Nested is the default. The encoding may be selected using one of the flags below: 

\begin{verbatim}
-smt-nested  (default)
-smt-flat
\end{verbatim}

The SMT backend is described and evaluated in a CP 2020 paper~\cite{davidson2020effective}, including comparison of the four theories and two encodings. 

Combined with \texttt{-run-solver}, the SMT backend will run Boolector, Z3, or Yices2. It will select a default solver for the chosen theory, or the solver can be set manually using one of the following flags:

\begin{verbatim}
-boolector  (default for QF_BV theory)
-z3         (default for QF_NIA)
-yices2     (default for QF_IDL and QF_LIA)
\end{verbatim}

Choosing a solver does not affect any other setting of the SMT backend. If the solver does not support the chosen theory (at the time of release) \savilerow will output a warning and continue. 

\subsubsection{Warning Flags}

The following flag produces a warning when the semantics of a model is affected by undefinedness (see \Cref{sec:undef}). 

\begin{verbatim}
-Wundef
\end{verbatim}

\subsubsection{Controlling \savilerow}

The following flag specifies a time limit in seconds. \savilerow will stop 
when the time limit is reached, unless it has completed tailoring and is 
running a backend solver. The time measured is wallclock time not 
CPU time. 

\begin{verbatim}
-timelimit <time>
\end{verbatim}

Time limits may also be passed through to solvers, for example when using Minion 
as the backend solver,
the following flag will limit Minion's run time (specified in seconds). 

\begin{verbatim}
-solver-options "-cpulimit <time>"
\end{verbatim}

A similar approach can be used to apply a time limit to other solvers. 

In some cases the SAT or MaxSAT output can be very large. The following
option allows the number of clauses to be limited. If the specified number of 
clauses is reached, \savilerow will delete the partial SAT file and exit. 

\begin{verbatim}
-cnflimit <numclauses>
\end{verbatim}

Some reformulations use a pseudorandom number generator with a fixed seed. The following option may be used to set a different seed. 

\begin{verbatim}
-seed <integer>
\end{verbatim}

\subsubsection{Solver Control}

The following flag causes \savilerow to run a solver and parse the solutions produced by it. 
This is currently implemented for Minion, Gecode, Chuffed, Standard FlatZinc, SAT, SMT, and MaxSAT backends. 

\begin{verbatim}
-run-solver
\end{verbatim}

The following two flags control the number of solutions. 
The first causes the solver to search for
all solutions (and \savilerow to parse all solutions). The second specifies a 
required number of solutions. 

\begin{verbatim}
-all-solutions
-num-solutions <n>
\end{verbatim}

When parsing solutions the default behaviour is to create one file 
for each solution. As an alternative, the following flag will send solutions 
to standard out, separated by lines of 10 minus signs. 

\begin{verbatim}
-solutions-to-stdout
\end{verbatim}

To prevent solutions being output at all, use the following flag. 

\begin{verbatim}
-solutions-to-null
\end{verbatim}

The following flag passes through command-line options to the solver. The string
would normally be quoted. 

\begin{verbatim}
-solver-options <string>
\end{verbatim}

For example when using Minion \texttt{-solver-options "-cpulimit <time>"} may
be used to impose a time limit, or when using Gecode \texttt{-solver-options "-p 8"}
causes Gecode to parallelise to 8 cores. 

\subsubsection{Solver Control -- Minion}

The following flag specifies where the Minion executable is. The default value
is \texttt{minion}. 

\begin{verbatim}
-minion-bin <filename>
\end{verbatim}

When Minion is run directly by \savilerow, preprocessing is usually 
applied before search starts. The following flag allows the level of 
preprocessing to be specified. 

\begin{verbatim}
-preprocess LEVEL 
    where LEVEL is one of None, GAC, SAC, SAC_limit, SSAC, SSAC_limit,
    SACBounds, SACBounds_limit, SSACBounds, SSACBounds_limit
\end{verbatim}

The default level of preprocessing is \texttt{SACBounds\textunderscore{}limit}.

The \texttt{-preprocess} flag affects the behaviour of Minion in two cases: first 
when Minion is called to filter domains (the \texttt{-reduce-domains} option), and 
second when Minion is called to search for a solution. 

\subsubsection{Solver Control -- Gecode}

The following flag specifies where the Gecode executable is. The default value
is \texttt{fzn-gecode}. 

\begin{verbatim}
-gecode-bin <filename>
\end{verbatim}

\subsubsection{Solver Control -- Chuffed}

The following flag specifies where the Chuffed executable is. The default value
is \texttt{fzn-chuffed}. 

\begin{verbatim}
-chuffed-bin <filename>
\end{verbatim}

\subsubsection{Solver Control -- OR Tools}

The following flag specifies where the OR Tools executable is. The default value
is \texttt{fzn-ortools}. 

\begin{verbatim}
-or-tools-bin <filename>
\end{verbatim}

By default OR Tools is called without its ``free search" option. To enable free search, use the pass-through flag:

\begin{verbatim}
-solver-options "-f"
\end{verbatim}

\subsubsection{Solver Control -- Standard FlatZinc}

The following flag specifies a FlatZinc solver binary. 

\begin{verbatim}
-fzn-bin <filename>
\end{verbatim}

\subsubsection{Solver Control -- SAT Solvers}

\savilerow is able to run and parse the output of MiniSAT, Lingeling, Glucose, CaDiCaL, and kissat solvers as well as two MiniSAT-based solvers for finding all solutions. The 
following flag specifies which family of solvers is used. Kissat is the default. 

\begin{verbatim}
-sat-family [ minisat | lingeling | glucose | cadical | kissat | 
              nbc_minisat_all | bc_minisat_all ]
\end{verbatim}

The `all' values imply the \texttt{-all-solutions} flag. 

The following flag specifies the SAT solver executable. 
The default value
is \texttt{minisat}, \texttt{lingeling}, \texttt{glucose}, \texttt{cadical}, \texttt{kissat}, \texttt{nbc\textunderscore{}minisat\textunderscore{}all\textunderscore{}release}, or \texttt{bc\textunderscore{}minisat\textunderscore{}all\textunderscore{}release} depending on the SAT family. 

\begin{verbatim}
-satsolver-bin <filename>
\end{verbatim}

The following flag enables interactive usage of the supported 
solvers when \savilerow is built with this 
feature. Supported SAT solvers are used 
incrementally via JNI calls. Currently 
supported: \texttt{glucose}, \texttt{cadical}, and 
\texttt{nbc\textunderscore{}minisat\textunderscore{}all}.

\begin{verbatim}
-interactive-sat
\end{verbatim}

When using SAT or SMT solvers, optimisation is performed with multiple calls to the solver. There are three modes: \textit{bisect} (default), \textit{linear}, and
\textit{unsat}. Each places constraints on the (encoding of the) objective variable.  
Bisect splits the interval of the objective variable into two approximately equal sub-intervals, and attempts to find a solution in the better sub-interval (i.e.\ smaller if minimising). The interval is updated and the procedure is repeated until the interval becomes empty, at which point the last solution found is an optimal solution. Linear bounds the objective variable each time a solution is found, such that the next solution will be better. Unsat assigns the objective variable to its best remaining value (smallest value when minimising). If no solution is found, the best value is removed and the procedure iterates, continuing until a solution is found. 

The optimisation strategy may be set with the following flag:

\begin{verbatim}
-opt-strategy [ bisect | linear | unsat ]
\end{verbatim}

\subsubsection{Solver Control -- SMT Solvers}

To set the SMT solver executable use the following flag. If it is not set, then \savilerow will look for a binary named \texttt{z3}, \texttt{yices-smt2}, or \texttt{boolector} in the bin directory (depending which SMT solver was selected, \texttt{boolector} by default, see \Cref{sec:smt-options}). 

\begin{verbatim}
-smtsolver-bin <filename>
\end{verbatim}

Note that the \texttt{-opt-strategy} flag is relevant when doing optimisation
with an SMT solver. 

\subsubsection{ReadSolution Mode}

\savilerow has two modes of operation, Normal and ReadSolution.  Normal is the
default, and in this mode \savilerow reads an \eprime model file and optional
parameter file and produces output for a solver. In some cases it will also 
run a solver and parse the solution(s), producing \eprime solution files. 

When using ReadSolution mode, \savilerow will read a solution in Minion
format and translate it back to \eprime. This allows the user to run Minion
separately from \savilerow but still retrieve the solution in \eprime format. 
When running Minion, the \texttt{-solsout} flag should be used to retrieve the
solution(s) in a table format. 

The mode is specified as follows.

\begin{verbatim}
-mode [Normal | ReadSolution]
\end{verbatim}

When using ReadSolution mode, the name of the aux file previously 
generated by \savilerow needs to be specified using the \texttt{-out-aux} flag,
so that \savilerow can read its symbol table. Also the name of the solution table file
from Minion is specified using \texttt{-minion-sol-file}. \texttt{-out-solution} 
is required to specify where to write the solutions. \texttt{-all-solutions} and 
\texttt{-num-solutions <n>} are optional in ReadSolution mode. 
Below is a typical command. 

\begin{verbatim}
savilerow -mode ReadSolution -out-aux <filename> -minion-sol-file <filename> 
                             -out-solution <filename>
\end{verbatim}

When neither \texttt{-all-solutions} nor \texttt{-num-solutions <n>} is given,
\savilerow parses the last solution in the Minion solutions file. For optimisation 
problems, the last solution will be the optimal or closest to optimal.

\appendix

\section*{Appendices}

\section{SAT Encoding} \label{app:satenc}

We have used standard encodings from the literature such as the order encoding for sums~\cite{tamura2009compiling} and support encoding~\cite{gent-encodings-02} for binary constraints. Also we do not attempt to encode all constraints in the language: several constraint types are decomposed before encoding to SAT. 

\subsection{Decomposition}

The first step is decomposition of the constraints AllDifferent, GCC, Atmost and Atleast. All are decomposed into sums of equalities and we have one sum for each relevant domain value. For example to decompose AllDifferent($[x,y,z]$): for each relevant domain value $a$ we have $(x=a)+(y=a)+(z=a)\le 1$ (or $(x=a)+(y=a)+(z=a) = 1$ when the number of variables and values are equal).  These decompositions are done after AC-CSE if AC-CSE is enabled (because the large number of sums generated hinders the AC-CSE algorithm) and before Identical and Active CSE. 

The second step is decomposition of lexicographic ordering constraints. We use the decomposition of Frisch et al~\cite{frisch:GACLex} (Sec.4) with implication rewritten as disjunction, thus the conjunctions of equalities in Frisch et al become disjunctions of disequalities. Also the common subsets of disequalities are factored out and replaced with auxiliary Boolean variables. 
This decomposition is also performed after AC-CSE and before Identical and Active CSE. 

The third step occurs after all flattening is completed. The constraints min, max, and element are decomposed. For $\mathrm{min}(V)=z$ we have $V[1]=z \vee V[2]=z \ldots$ and $z\le V[1] \wedge z\le V[2] \ldots$. Max is similar to min with $\le$ replaced by $\ge$. The constraint $\mathrm{element}(V, x)=z$ becomes $\forall i : (x\ne i \vee V[i]=z)$. 

\subsection{Encoding of CSP variables}

The encoding of a CSP variable provides SAT literals for facts about the variable: $[x=a]$, $[x\ne a]$, $[x\le a]$ and $[x>a]$ for a CSP variable $x$ and value $a$. We will refer to $[x=a]$ and $[x\ne a]$ as direct literals, and $[x\le a]$ and $[x>a]$ as order literals. 
If the variable has only two values, it is represented with a single SAT variable. All the above facts (for both domain values) map to the SAT variable, its negation, \textit{true} or \textit{false}. 

When the CSP variable has more than two values, an encoding is chosen to provide the literals that are needed to encode the constraints. There are three encodings, with one providing the order literals, another providing the direct literals, and the third providing both. Encodings of the binary constraints \(<\) and \(\leq\) use the order literals, and the tree encoding (see Section~\ref{sec:sat-options}) of sums also uses the order literals. All other constraint encodings use the direct literals. If the objective function is a single variable, the order literals are required for that variable. Otherwise, if the objective function is a sum at encoding time (only possible with the MaxSAT backend) then direct literals are required for all variables in the sum. 

If the order literals are not needed for a variable \(x\), then the variable is encoded with one SAT variable per domain value (representing \([x=a]\)). The 2-product encoding is used for at-most-one and a single clause for at-least-one. 

If the direct literals are not needed, then using the language of the \textit{ladder} encoding of Gent et al~\cite{GentIP:alldiff-to-sat}, we have only the ladder variables and the clauses in Gent et al formula (2). If both order and direct literals are required, then we use the full ladder encoding with the clauses in formulas (1), (2) and (3) of Gent et al. Also, for the maximum value $\mathrm{max}(D(x))$ the facts $[x\ne \mathrm{max}(D(x))]$ and $[x<\mathrm{max}(D(x))]$ are equivalent so one SAT variable is saved. For the minimum value $\mathrm{min}(D(x))$, the facts $[x=\mathrm{min}(D(x))]$ and $[x\leq\mathrm{min}(D(x))]$ are equivalent so one SAT variable is saved.  Also, a variable may have gaps in its domain. Suppose variable $x$ has domain $D(x)=\{1\ldots 3, 8\ldots 10\}$. SAT variables are created only for the 6 values in the domain. Facts containing values $\{4\ldots 7\}$ are mapped appropriately (for example $[x\le 5]$ is mapped to $[x\le 3]$).
The encoding has $2|D(x)|-3$ SAT variables. 

\subsection{Encoding of Constraints}

Now we turn to encoding of constraints. Some simple expressions such as $x=a$, $x\le a$ and $\neg x$ (for CSP variable $x$ and value $a$) are represented with a single SAT literal. We have introduced a new expression type named SATLiteral. Each expression that can be represented as a single literal is replaced with a SATLiteral in a final rewriting pass before encoding constraints. SATLiterals behave like boolean variables hence they can be transparently included in any constraint expression that takes a boolean subexpression. 

At-most-one and exactly-one constraints (i.e.\ a sum of boolean expressions that is $\leq 1$ or $=1$) are by default encoded with the 2-product encoding, which is small and tends to perform better than alternatives. Other encodings can be selected by command-line flags (Section~\ref{sec:sat-options}). 

There is a choice of 5 encodings of pseudo-Boolean and other sum constraints, as described in Section~\ref{sec:sat-options}. The MDD, GPW, SWC, and GGT encodings are implemented exactly as described in Bofill et al \cite{amo-pb-cpaior19}. Sum-equal constraints are split into sum-$\le$ and sum-$\ge$ before encoding. All four of these encodings work with automatic AMO and EO detection as described in the CP 2019 paper~\cite{amopb-cp19}. The flag \texttt{-amo-detect} is used to switch on the AMO and EO detection. 

The Tree encoding for sums decomposes each sum constraint using a binary tree, similar to the generalized totalizer. Auxiliary integer variables are introduced for the internal nodes of the tree. A ternary sum constraint is used to connect each internal node to its two children. The ternary sums are then encoded using the order encoding~\cite{tamura2009compiling}. The Tree encoding does not work with AMO and EO detection. Tree can be superior to MDD, GPW, SWC, and GGT in the case where the original constraint is a sum-equality, because Tree does not first break down the sum-equality constraint into sum-$\le$ and sum-$\ge$. 

For other constraints we used the standard support encoding wherever possible~\cite{gent-encodings-02}. Binary constraints such as $|x|=y$ use the support encoding, and ternary functional constraints $x\diamond y=z$ (eg $x\times y=z$) use the support encoding when $z$ is a constant. Otherwise, $x\diamond y=z$ are encoded as a set of ternary SAT clauses: $\forall i \in D(x), \forall j \in D(y): (x\ne i \: \vee\: y\ne j \: \vee\: z=i\diamond j)$. When $i\diamond j$ is not in the domain of $z$, the literal $z=i\diamond j$ will be false. 
Logical connectives such as $\wedge, \vee, \leftrightarrow$ have custom encodings. Binary table constraints use the support encoding. Non-binary tables use Bacchus' encoding~\cite{bacchus2007gac} (Sec.2.1) by default. An MDD-based encoding is also implemented for non-binary table constraints and is enabled with a command-line flag (\Cref{sec:sat-options}).


\section{Operator Precedence in \eprime}\label{app:op}

Table \ref{tab:precedence} shows the precedence and associativity of operators
in \eprime. As you would expect, operators with higher precedence are applied first.

Left-associative operators are evaluated left-first, for example {\tt 2/3/4 = (2/3)/4}.
The only operator with right associativity is {\tt **}. This allows double
exponentiation to have its conventional meaning: {\tt 2**3**4 = 2**(3**4)}

Unary operators usually have a higher precedence than binary ones. There is one
exception to this rule: that \texttt{**} has a higher precedence than unary minus. 
This allows \texttt{-2**2**3} to have its conventional meaning 
of \texttt{-(2**(2**3))=-256}, as opposed to \texttt{(-2)**(2**3)=256}.

\begin{table}
\begin{center}
\begin{tabular}{llcc}
\toprule
Operator & Functionality & Associativity & Precedence \\
\hline
{\tt !} & Boolean negation &  & 20\\
{\tt ||} & Absolute value  &  & 20\\
\hline
{\tt **} & Power & Right & 18 \\
\hline
{\tt -} & Integer negation &   & 15\\
\hline
{\tt *} & Multiplication & Left & 10 \\
{\tt /} & Division & Left & 10 \\
{\tt \%} & Modulo & Left & 10 \\
\hline
{\tt intersect} & Domain intersection & Left & 2 \\
\hline
{\tt union} & Domain union & Left & 1 \\
{\tt +} & Addition & Left & 1 \\
{\tt -} & Subtraction \& Domain Subtraction & Left & 1 \\
\hline
{\tt =} & Equality & None & 0 \\
{\tt !=} & Disequality & None & 0 \\
{\tt <=} & Less-equal & None & 0 \\
{\tt <} & Less than & None & 0 \\
{\tt >=} & Greater-equal & None & 0 \\
{\tt >} & Greater than & None & 0 \\
{\tt <=lex} & Lex less-equal & Left & 0 \\
{\tt <lex} & Lex less than & Left & 0 \\
{\tt >=lex} & Lex greater-equal & Left & 0 \\
{\tt >lex} & Lex greater than & Left & 0 \\
\hline
{\tt in} & Value in a set & Left & 0 \\
\hline
{\tt \verb1/\1} & And & Left & -1 \\
\hline
{\tt \verb1\/1} & Or & Left & -2 \\
\hline
{\tt ->} & Implication & None & -4 \\
{\tt <->} & If and only if & None & -4 \\
\hline
{\tt forAll, exists, sum} & Quantifiers & & -10 \\
\hline
{\tt ,} & And & Left & -20 \\
\bottomrule
      \end{tabular}
\end{center}
    \caption{Operator precedence in \eprime}

\label{tab:precedence}
\end{table}

\section{Reserved Words} \label{app:reservedwords}

The following words are keywords and therefore are not allowed to be used as identifiers. 

\begin{verbatim}
forall, forAll, exists, sum,  
such, that, letting, given, where, find, language, 
int, bool, union, intersect, in, false, true
\end{verbatim}

\bibliographystyle{plain}
\bibliography{general}

\end{document}